\documentclass[letterpaper,10pt,conference]{ieeeconf}

\IEEEoverridecommandlockouts
\usepackage{amsmath,amssymb,amsfonts}
\usepackage{graphicx}
\usepackage{booktabs}
\usepackage{flafter}
\usepackage{cite}
\usepackage{url}
\usepackage[caption=false,font=footnotesize,position=bottom]{subfig}
\usepackage[hidelinks,bookmarks=false]{hyperref}

\graphicspath{{figures/}}

\newcommand{\bx}{\mathbf{x}}
\newcommand{\bu}{\mathbf{u}}
\newcommand{\bp}{\mathbf{p}}
\newcommand{\bv}{\mathbf{v}}

\newcommand{\be}{\mathbf{e}}
\newcommand{\bmu}{\boldsymbol{\mu}}
\newcommand{\R}{\mathbb{R}}

\newcommand{\norm}[1]{\left\|#1\right\|}
\begin{document}

\title{\LARGE\bfseries A Game-Theoretic MPC Planner for Cooperative and Adversarial Drone Coordination}

\author{Zden\v{e}k Bou\v{c}ek and Miroslav Fl\'{i}dr%
\thanks{Z. Bou\v{c}ek and M. Fl\'{i}dr are with the New Technologies for the
Information Society Research Center \& Department of Cybernetics, Faculty of
Applied Sciences, University of West Bohemia in Pilsen, Czechia. E-mail:
\texttt{\{zboucek,flidr\}@fav.zcu.cz}.}}

\maketitle

\begin{abstract}
We present and evaluate a game-theoretic model predictive control planner for cooperative goal reaching and adversarial pursuit--evasion with drones. The planner combines 3D prediction, individual objectives, optional shared separation constraints and trajectory-tracking interfaces. A comparison of cooperative game solvers assesses their ability to find feasible trajectories that resolve conflicts. A computational study quantifies how number of drones, prediction horizon and move blocking affect computation time and numerical acceptance. Move blocking holds control inputs constant over groups of prediction steps, reducing the number of independent control variables and lowering computation time. Timing measurements identify the practical limits of real-time planning in the tested configurations. PyBullet simulations demonstrate cooperative goal reaching and pursuit--evasion. A laboratory flight complements the simulations by validating the trajectory-tracking interfaces on physical drones. The planner can serve as a basis for further evaluation and controller development.
\end{abstract}


\section{Introduction}
\label{sec:intro}

Coordinating multiple drones requires planning trajectories that account for both vehicle dynamics and the motion of other drones. Cooperative inspection and environmental monitoring motivate such planning in shared airspace \cite{chung2018survey,du2025survey}. In cooperative goal reaching, each drone seeks its assigned position while maintaining separation from the others \cite{nascimento2023safe}. In adversarial pursuit--evasion, the drones have opposing objectives: pursuers seek to approach evaders, while evaders seek to avoid them \cite{hoogendoorn2026conflict}.

Model predictive control (MPC) provides a way to plan motion using a dynamics model and state and input constraints \cite{mayne2000constrained,nguyen2021mpcsurvey}. It optimizes future control inputs over a finite horizon and repeats the optimization after executing the first part of the plan. In this work, MPC serves as a trajectory-planning layer that supplies predicted positions and velocities to a tracking controller.

Each drone's preferred motion depends on how the other drones move. This interaction can be modeled as a game in which each drone chooses a control sequence to minimize its own cost \cite{fridovichkeil2020ilqgames}. In game-theoretic MPC, this game is solved over the prediction horizon at each planning update \cite{liu2024stability}.

A Nash equilibrium (NE) is a set of plans in which each drone's plan is optimal given the plans of the others \cite{rosen1965existence}. If minimum separation is enforced through shared constraints, each drone's feasible plans also depend on the other drones' trajectories. This leads to a generalized Nash equilibrium (GNE) problem \cite{facchinei2007generalized}.

Implementing this game-theoretic MPC planner requires assessing both the computed plans and the time needed to obtain them. The interaction costs and separation constraints can make the game nonconvex. A numerical solver may then return a candidate that satisfies necessary optimality conditions without being an equilibrium \cite{bemporad2025nashopt}. We also assess whether the computed plans resolve conflicts. Computational cost depends on the number of drones, prediction horizon and control parameterization. We therefore investigate how to implement the planner, how to evaluate its solutions and whether their computation fits the available planning period.

We address these questions by implementing and evaluating a game-theoretic MPC planner for cooperative and adversarial drone coordination. Our contributions are:
\begin{enumerate}
\item A reusable game-theoretic MPC planner for 3D drone coordination, accompanied by executable cooperative and adversarial scenarios.
\item A numerical comparison of joint equilibrium searches and exact-potential minimization, assessing numerical acceptance, unilateral improvements and conflict resolution. Flight simulations evaluate closed-loop behavior.
\item A computational study quantifying how number of drones, prediction horizon and move blocking affect computation time, constraint satisfaction and numerical acceptance.
\end{enumerate}

A five-drone laboratory flight additionally tests the tracking interfaces.

The remainder of the paper is organized as follows. Section~\ref{sec:related} reviews game-theoretic drone planning and numerical game solvers. Section~\ref{sec:problem} presents the drone models and game formulation. Section~\ref{sec:mpc} describes objective function design, numerical game solution and receding-horizon implementation. Section~\ref{sec:experiments} evaluates numerical solution reliability and computation. Section~\ref{sec:flight} presents flight simulations and laboratory tracking integration. Section~\ref{sec:conclusion} concludes the paper and outlines future work.

\section{Related Work}
\label{sec:related}

Game-theoretic drone planning already covers both cooperation and conflict. In competitive drone racing, Wang et al. \cite{wang2020seibr} use shared collision constraints and sensitivity-enhanced iterative best response, including flight experiments. For cooperative coordination, Shi et al. \cite{shi2022potential} use generalized potential games and sequential drone responses, while do Nascimento et al. \cite{nascimento2023safe} combine GNE-MPC with control barrier functions for distributed drone control. Hoogendoorn et al. \cite{hoogendoorn2026conflict} describe 3D conflict resolution and pursuit--evasion within a receding-horizon Nash framework using soft proximity penalties. These studies establish the planning approach. They also motivate distinguishing hard shared constraints from interactions represented only in the costs.

Numerical methods address different parts of this planning problem. iLQGames \cite{fridovichkeil2020ilqgames} uses iterative linear-quadratic approximations for nonlinear multiplayer games. ALGAMES \cite{lecleach2022algames} handles constrained dynamic games through an augmented-Lagrangian method and a common shared-constraint multiplier formulation. NashOpt \cite{bemporad2025nashopt} provides joint optimality-condition solvers for standard and variational GNE formulations. Hall and Bemporad \cite{hall2025multiparametric} derive explicit equilibrium maps for quadratic games with linear constraints, while Liu and Kolmanovsky \cite{liu2024stability} study Newton methods for computing and tracking equilibria in MPC.

We complement the cooperative coordination studies of Shi et al. and do Nascimento et al.~\cite{shi2022potential,nascimento2023safe} by comparing standard and variational GNE candidates with exact-potential minimization under matched cooperative costs. We relate numerical acceptance and feasible unilateral cost improvements to conflict resolution, following the response checks recommended for NashOpt~\cite{bemporad2025nashopt}. Our cooperative game includes hard shared separation constraints, whereas Hoogendoorn et al.~\cite{hoogendoorn2026conflict} represent proximity through soft costs. Flight simulations and matched computational measurements assess task completion and computation time.

\section{Problem Formulation}
\label{sec:problem}

We formulate drone coordination as a finite-horizon game in which each drone chooses an acceleration sequence to minimize its own cost. Each drone corresponds to one player in the game. The planner assumes known current states, nominal dynamics and objectives for all $N$ drones. We first define the prediction model and motion constraints, then introduce the game and its equilibrium concepts.

\subsection{Drone Dynamics and Constraints}

The $i$-th drone, $i \in \{1,\ldots,N\}$, is modeled as a 3D double integrator with position/velocity state $\bx_{i,k} = [p_x, p_y, p_z, v_x, v_y, v_z]^\top \in \R^6$ and desired acceleration $\bu_{i,k} = [a_x, a_y, a_z]^\top \in \R^3$. With time step $\Delta t$, its dynamics is
\begin{equation}
  \bx_{i,k+1} = A\,\bx_{i,k} + B\,\bu_{i,k},
  \label{eq:dynamics}
\end{equation}
where $A=\left[\begin{smallmatrix}I_3 & \Delta t\,I_3 \\ 0_{3\times3} & I_3\end{smallmatrix}\right]$ and $B=\left[\begin{smallmatrix}\tfrac{1}{2}\Delta t^2 I_3 \\ \Delta t\,I_3\end{smallmatrix}\right]$.

Predictions over $H$ steps follow \eqref{eq:dynamics} given the initial state $\bx_{i,0}$. Vectors $\bp_{i,k}$ and $\bv_{i,k}$ denote the position and velocity components.

Local constraints can bound acceleration, altitude, speed and acceleration increments. Define $\Delta\bu_{i,k}=\bu_{i,k}-\bu_{i,k-1}$, where $\bu_{i,-1}$ is the known input from the preceding control update. The bounds take the form
\begin{equation}
  \begin{aligned}
  \norm{\bu_{i,k}}_\infty&\le a_{\max,i},\\
  z_{\min,i}&\le p_{z,i,k+1}\le z_{\max,i},\\
  \norm{\bv_{i,k+1}}_2&\le v_{\max,i},\\
  \norm{\Delta\bu_{i,k}}_\infty&\le\Delta t\,j_{\max,i},
  \end{aligned}
  \label{eq:local_bounds}
\end{equation}
for $k=0,\ldots,H-1$. Acceleration and discrete jerk $\Delta\bu_{i,k}/\Delta t$ are bounded per axis.

\subsection{Finite-Horizon Game and Equilibrium Concepts}

We stack the control sequences and states as follows, where $-i$ denotes the exclusion of the $i$-th drone:
\begin{equation*}
  \begin{aligned}
  \bu_i &= [\bu_{i,0}^\top,\ldots,\bu_{i,H-1}^\top]^\top, \\
  \bu &= [\bu_1^\top,\ldots,\bu_N^\top]^\top, \\
  \bu_{-i} &= [\bu_1^\top,\ldots,\bu_{i-1}^\top,\bu_{i+1}^\top,\ldots,\bu_N^\top]^\top, \\
  \bx_{-i,k} &= [\bx_{1,k}^\top,\ldots,\bx_{i-1,k}^\top,\bx_{i+1,k}^\top,\ldots,\bx_{N,k}^\top]^\top.
  \end{aligned}
\end{equation*}

Let $\bx_0=[\bx_{1,0}^\top,\ldots,\bx_{N,0}^\top]^\top$ denote the joint initial state and $J_i(\bu_i,\bu_{-i};\bx_0)$ the $i$-th drone's scenario-dependent cost, specified in Section~\ref{sec:objective_design}.
With the initial state $\bx_0$ as a fixed parameter and the opponents' control sequences $\bu_{-i}$ held fixed, the $i$-th drone solves
\begin{equation}
  \min_{\bu_i\in\mathcal{F}_i(\bu_{-i};\bx_0)}
  J_i(\bu_i,\bu_{-i};\bx_0).
  \label{eq:player_problem}
\end{equation}
The feasible-response set $\mathcal{F}_i(\bu_{-i};\bx_0)$ contains control sequences satisfying the enabled local bounds \eqref{eq:local_bounds} and any enabled shared separation constraints below.

Shared separation constraints, if enabled, are defined as
\begin{equation}
  g^{\mathrm{col}}_{ij,k}(\bu;\bx_0) = d_{\min}^2
  - \norm{\bp_{i,k+1}-\bp_{j,k+1}}_2^2 \le 0,
  \label{eq:collision}
\end{equation}
for all pairs $i<j$ and prediction steps $k=0,\ldots,H-1$, where $d_{\min}$ is the required minimum separation. When shared separation constraints are disabled, each drone's feasible set is independent of the other drones' plans, giving an ordinary NE problem.

A feasible joint control vector $\bu$ is a GNE if each drone's sequence $\bu_i$ solves \eqref{eq:player_problem} with the other sequences $\bu_{-i}$ held fixed \cite{facchinei2007generalized}. Equivalently,
\begin{equation}
  \begin{aligned}
  J_i(\bu_i,\bu_{-i};\bx_0)\le J_i(\tilde{\bu}_i,\bu_{-i};\bx_0),\\
  \forall\,\tilde{\bu}_i\in\mathcal{F}_i(\bu_{-i};\bx_0),\;\forall\,i,
  \end{aligned}
  \label{eq:gne}
\end{equation}
where $\tilde{\bu}_i$ denotes an alternative feasible control sequence for the $i$-th drone. Thus, no drone can reduce its cost by changing its own sequence alone.

The Lagrange multipliers associated with the shared constraints are denoted by $\bmu_i$ for the $i$-th drone. They describe how small changes in the constraint bounds affect the $i$-th drone's optimal cost. A standard GNE allows these multiplier vectors to differ across drones. A special case imposes a common vector, $\bmu_1=\cdots=\bmu_N=\bmu$. Under suitable convexity and regularity assumptions, equilibria satisfying this condition are called \emph{variational} GNE (vGNE) \cite{rosen1965existence,facchinei2007generalized}. In our nonconvex games, the common-multiplier condition is used to compute vGNE candidates and does not by itself establish an equilibrium.

\section{MPC Design and Implementation}
\label{sec:mpc}

The finite-horizon game needs to be embedded in an MPC planner that supplies predicted positions and velocities to a tracking controller. We specify the objective functions for cooperative goal reaching and adversarial pursuit--evasion, followed by the numerical methods used to compute candidate solutions. The receding-horizon implementation uses warm starts and move blocking to support repeated optimization.

\subsection{Objective Function Design}
\label{sec:objective_design}

Let $\be_{i,k}=\bx_{i,k}-\bx_{i,\mathrm{ref}}$, where $\bx_{i,\mathrm{ref}}=[\bp_{i,\mathrm{ref}}^\top,0,0,0]^\top$. The $i$-th drone's scenario-dependent cost is
\begin{equation}
  \begin{aligned}
  &J_i(\bu_i,\bu_{-i};\bx_0) = \sum_{k=1}^{H-1}
    \be_{i,k}^\top Q_i\be_{i,k} \\
  &\quad{}+\sum_{k=0}^{H-1}\!\Bigl(\bu_{i,k}^\top R_i\bu_{i,k}
    + \Delta\bu_{i,k}^\top R_{\Delta,i}\Delta\bu_{i,k} \\
  &\qquad{}+ \Phi_i(\bx_{i,k+1},\bx_{-i,k+1})\Bigr)
    + \be_{i,H}^\top Q_{F,i}\be_{i,H},
  \end{aligned}
  \label{eq:cost}
\end{equation}
where the matrix $Q_i=\operatorname{diag}(Q_{p,i},Q_{v,i})$ weights intermediate position/velocity errors, and $Q_{F,i}$ weights final state error. Matrices $R_i$ and $R_{\Delta,i}$ weight acceleration and input increments. All weight matrices are symmetric, with $Q_i,Q_{F,i},R_i,R_{\Delta,i}\succeq0$ for all $i$. We additionally assume
\begin{equation*}
  \forall i\in\{1,\ldots,N\}:\quad
  (R_i\succ0)\lor(R_{\Delta,i}\succ0).
\end{equation*}
With $\bu_{i,-1}$ fixed, this is a sufficient condition for the control penalty to grow without bound as $\norm{\bu_i}_2\to\infty$. Input-increment bounds and penalties are enabled independently. The term $\Phi_i$ contains role-dependent interactions and soft obstacle penalties.

\paragraph*{Cooperative goal reaching}
Cooperative drones track assigned target positions and penalize pairwise proximity through $\Phi_i$ in \eqref{eq:cost}.

In cooperative games without obstacles, both drones' objectives contain the same pairwise repulsion cost, defined for \mbox{$1\leq i<j\leq N$} as
\begin{equation}
 C_{ij}(\bu_i,\bu_j;\bx_0)
 =w_{\mathrm{rep}}\sum_{k=0}^{H-1}
 \frac{1}{\norm{\bp_{i,k+1}-\bp_{j,k+1}}_2^2+\varepsilon}.
\end{equation}
The common weight $w_{\mathrm{rep}}\geq0$ sets repulsion strength, and $\varepsilon=0.01\,\mathrm{m}^2$ keeps the penalty finite at zero distance. Hard separation is imposed separately by \eqref{eq:collision}.

\paragraph*{Pursuit--evasion}
In pursuit--evasion, pursuers penalize squared nearest-evader distance and omit own-state tracking and terminal penalties ($Q_i=0$, $Q_{F,i}=0$). Evaders penalize target error and inverse distance to the nearest pursuer, with additional velocity and terminal penalties in flight. Both roles penalize control effort and use soft repulsion between drones. Selecting the nearest opponent by an exact minimum makes these interactions generally nonsmooth at ties.

\subsection{Numerical Game Solution}
\label{sec:method}

For the cooperative game, $\sum_i J_i$ counts each pair twice, so the exact potential
\begin{equation}
 P(\bu;\bx_0)=\sum_iJ_i(\bu_i,\bu_{-i};\bx_0)
 -\sum_{i<j}C_{ij}(\bu_i,\bu_j;\bx_0)
 \label{eq:potential}
\end{equation}
subtracts one copy of each pair contribution. For fixed $\bx_0$ and opponents' plans, a change in $\bu_i$ changes $P$ and $J_i$ equally. A feasible global minimizer therefore admits no feasible improving unilateral response. We call joint minimization of $P$ over all drones' control sequences, subject to the local bounds and shared separation constraints, \emph{potential minimization}. Numerical solvers seek local minima, so their candidates require separate equilibrium-quality checks.

For cooperative flight, GNE candidates are computed by sequential minimization of each drone's cost and vGNE candidates by joint minimization of $P$. The latter uses common shared-constraint multipliers. Active-set refinement updates controls and multipliers, and feasible improving unilateral responses restart the search. Joint Karush--Kuhn--Tucker (KKT) conditions or local potential minimization alone do not establish global equilibrium in these nonconvex games \cite{bemporad2025nashopt}. Feasibility and equilibrium-quality diagnostics are reported in Sections~\ref{sec:selection} and~\ref{sec:cooperative_flight}.

Pursuit--evasion uses sequential response sweeps. In each sweep, drones update their control sequences one at a time, locally minimizing their own cost while the other sequences are fixed. Let $\widehat{\bu}_i$ be the locally optimized response and $\bu_i$ the current sequence. We test candidate updates $\bu_i+\alpha(\widehat{\bu}_i-\bu_i)$, where $0<\alpha\leq1$. Fractions below one damp the changes between successive player responses. The first feasible update that does not increase the drone's cost is accepted. From an infeasible initialization, restoring feasibility takes priority. Damping is a numerical heuristic and does not guarantee convergence of the response iterations. The final plan is checked for model feasibility, without a separate search for improving unilateral responses.

\subsection{Receding-Horizon Implementation}
\label{sec:blocked_method}
At each update, the planner computes acceleration sequences and passes predicted positions and velocities to the tracking controller. Only the first interval is executed.

Each subsequent optimization is warm started by shifting the previous control sequences one prediction step and appending zero acceleration. To reduce the number of independent control variables, we group consecutive prediction steps into blocks and hold the control inputs constant within each block. This parameterization is called move blocking \cite{cagienard2007blocking}. Costs and constraints are evaluated at every node. The shifted sequence is projected onto the block grid and checked for feasibility.

\section{Numerical Evaluation}
\label{sec:experiments}
\label{sec:experimental_setup}

We compare joint KKT searches and potential minimization for resolving conflicts between drones, then measure planning times with and without move blocking. All experiments in this section use 3D double-integrator dynamics with $\Delta t=0.1$\,s and run on an Intel Core i9-9900 CPU. The timing benchmark uses SciPy's SLSQP, a sequential quadratic programming solver. The nonlinear experiments set $R_{\Delta,i}=0$, while laboratory tracking uses input-increment penalties (Sec.~\ref{sec:hardware}).

\subsection{Solution Reliability and Conflict Resolution}
\label{sec:selection}
\label{sec:acceptance}

We compare joint KKT searches and potential minimization in a three-drone cooperative game with prediction horizon $H=5$, individual controls and hard 0.50\,m node separation. All methods share states, targets, costs and initial guesses. Tracking, effort and symmetric repulsion follow \eqref{eq:potential}, with $w_{\mathrm{rep}}=1$. Velocity and input-increment penalties are disabled, and $Q_{F,i}=Q_i$. This simplified game isolates numerical solution behavior from the longer horizon and move blocking used in flight. The joint KKT searches use NashOpt~1.1.0 \cite{bemporad2025nashopt} for standard GNE and vGNE, with individual and common shared-constraint multipliers, respectively. Potential minimization uses SLSQP to minimize $P$ as described in Section~\ref{sec:method}. Numerical acceptance checks feasibility, first-order residuals and multipliers.

\begin{figure*}[t]
\centering
\captionsetup[subfloat]{farskip=0pt}
\begin{minipage}[b]{0.67\textwidth}
\centering
\includegraphics[width=0.40\linewidth]{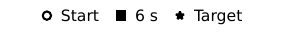}\par\nointerlineskip
\subfloat[Joint KKT (GNE), 0--6\,s\label{fig:numerical_conflict_gne}]{\includegraphics[width=0.49\linewidth]{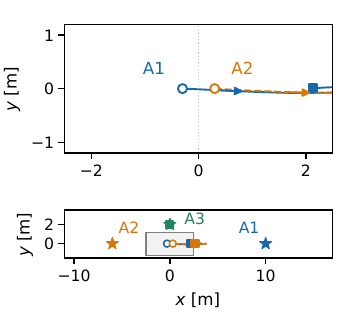}}\hfill
\subfloat[Potential minimization, 0--6\,s\label{fig:numerical_conflict_potential}]{\includegraphics[width=0.49\linewidth]{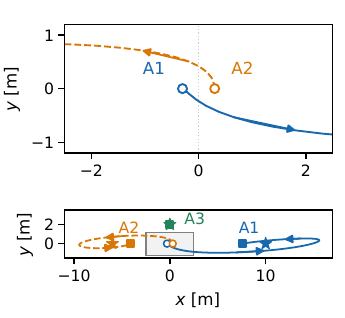}}
\end{minipage}\hfill
\begin{minipage}[b]{0.31\textwidth}
\centering
\subfloat[Conflict passage, 0--6\,s\label{fig:numerical_conflict_passage}]{\includegraphics[width=\linewidth]{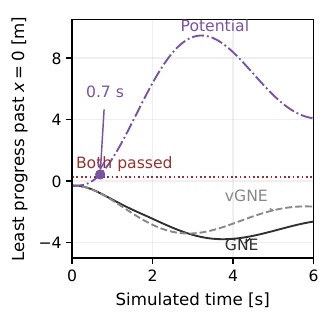}}
\end{minipage}
\caption{Head-on conflict with three drones. (a)--(b) show spatial details above complete paths, with matching spatial scales and boxes locating the details. Colors identify drones. Circles mark starts, squares positions at 6\,s, stars targets and arrows motion. Joint KKT trajectories are shown for GNE, while (c) includes GNE, vGNE and potential minimization over the same 0--6\,s window.}
\label{fig:numerical_conflict}
\end{figure*}

Figure~\ref{fig:numerical_conflict} illustrates a head-on conflict between two moving drones and a third whose target is its initial position. In Figure~\ref{fig:numerical_conflict_gne}, the GNE search initially moves both drones right, taking A2 away from its target. Potential minimization sends them past each other on opposite sides (Figure~\ref{fig:numerical_conflict_potential}). Both joint KKT methods accept all 60 updates but fail to pass the conflict. Potential minimization first achieves joint passage at 0.7\,s. The moving drones subsequently overshoot and reverse direction, but do not come to rest at their target positions within the 6\,s episode. Figure~\ref{fig:numerical_conflict_passage} compares all three methods over 0--6\,s using $\min(x_{\mathrm{A1}},-x_{\mathrm{A2}})$. Reaching 0.25\,m means both moving drones have passed the conflict center. The stationary-target drone is excluded from this measure.

We extend the comparison to orthogonal and three-way encounters. Three geometries, five state variants and six common initializations give 90 single-game optimization runs per method. Two initializations also generate 30 receding-horizon episodes per method, each requesting 60 updates and stopping on numerical rejection. Conflict passage requires every initially moving drone to be at least 0.25\,m beyond the origin along its route at a common instant.

\begin{table}[!htbp]
\caption{Equilibrium search outcomes}
\label{tab:controlled_results}
\centering\small
\setlength{\tabcolsep}{2pt}
\begin{tabular}{@{}lrrrr@{}}
\toprule
Method & \shortstack{Static\\accepted} & \shortstack{Episodes\\completed} & \shortstack{Conflict\\passed} & \shortstack{Improvement\\found} \\
\midrule
Joint KKT (GNE) & 42/90 & 10/30 & 4/30 & 49/59 \\
Joint KKT (vGNE) & 42/90 & 10/30 & 4/30 & 25/33 \\
Potential minimization & \textbf{90/90} & \textbf{30/30} & \textbf{30/30} & \textbf{0/90} \\
\bottomrule
\end{tabular}
\end{table}

Table~\ref{tab:controlled_results} separates numerical acceptance from task progress and equilibrium quality. The last column reports the number of candidates for which a feasible unilateral improvement is found, out of the number checked. These checks use distinct, numerically accepted candidates from single-game optimization runs and episode endpoints. A feasible improving unilateral response refutes an accepted equilibrium candidate. Multistart searches hold opponents fixed and test cost reductions above $\max(10^{-5},10^{-6}|J_i|)$. They refute many accepted joint KKT candidates, whereas no such improvement is found for candidates obtained by potential minimization. This method passes every tested conflict. Failure to find an improvement does not prove global equilibrium. These observations concern the tested configurations, initializations and search budgets. The short horizon may influence progress and candidate selection, but its effect is not isolated here.

\subsection{Computational Performance}
\label{sec:realtime}

\begin{figure}[t]
\centering
\includegraphics[width=\columnwidth]{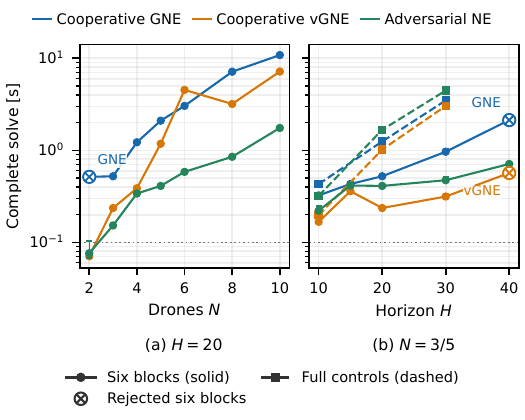}
\caption{Total computation times versus (a) drone count and (b) horizon. Solid circles use move blocking with six blocks, dashed squares use separate controls at each prediction step. Error bars span the minimum and maximum times across runs. Filled markers pass all five respective checks. Crossed hollow markers retain rejected candidates with adjacent method labels. Cooperative checks include local equilibrium tests, adversarial checks model feasibility. The dotted line marks 100\,ms.}
\label{fig:scalability}
\end{figure}

We measure whether planning fits within a 100\,ms period and how move blocking changes workload (Figure~\ref{fig:scalability}). For each configuration, one warm-up run is followed by five sequential runs at the same state, with the underlying linear algebra library restricted to a single CPU thread. The warm-up run is excluded from the timing statistics. Timing includes projection, optimization, polishing and numerical checks/repairs. Model preparation, physics, tracking control and supervisory retries are excluded. Matched cases with and without move blocking preserve costs, bounds, states and expanded initial controls. Move blocking restricts admissible plans, so timings do not imply identical equilibria. Repeated runs quantify timing variability at fixed inputs rather than success rates across encounters.

At $H=30$, move blocking with six blocks reduces median runtime by factors 3.61/9.65/9.43 for GNE/vGNE/adversarial responses. Flight-sized cases with move blocking at $H=20$ take 0.523/0.236\,s for three-drone GNE/vGNE and 0.411\,s for five-drone pursuit--evasion, all exceeding 100\,ms. Both cooperative $H=40$ cases fail numerical acceptance: GNE returns feasible approximations and vGNE infeasible candidates. GNE at $(N,H)=(2,20)$ also fails local checks. Cooperative checks assess feasibility and local equilibrium quality. All adversarial calls pass model-feasibility checks, without a separate equilibrium-quality test. Rejected candidates remain marked in the graph. Move blocking lowers computation time but does not ensure numerical acceptance or operation within the planning period.

\section{Flight Simulations and Laboratory Integration}
\label{sec:flight}

Flights use Crazyflie CF2X dynamics in PyBullet through gym-pybullet-drones \cite{panerati2021learning}, simulator state feedback and proportional--integral--derivative (PID) tracking of predicted positions/velocities. Physics, PID and planning run at 240/120/10\,Hz of simulated time. Simulation time pauses during optimization. Both scenarios use $H=20$, $\Delta t=0.1$\,s and move blocking (Section~\ref{sec:blocked_method}) with block lengths $(1,1,2,4,6,6)$ measured in prediction steps. Acceleration/speed/jerk limits are 2\,m/s$^2$, 2\,m/s and 6\,m/s$^3$, and heights lie in $[0.35,1.95]$\,m. Their longer horizon, move blocking and terminal weights do not allow a matched backend comparison with Section~\ref{sec:selection}.

\subsection{Cooperative Flight}
\label{sec:cooperative_flight}

Three drones combine head-on and transverse motion in obstacle-free space. SciPy's SLSQP performs the cooperative minimizations described in Section~\ref{sec:method}. Planned/monitored separation is 0.55/0.50\,m, distinguishing the planning reserve from measured clearance. Weights are $Q_i=I_6$, $R_i=0.1I_3$ and $R_{\Delta,i}=0$. The Riccati terminal matrix is computed at configuration. If the solver does not find a feasible plan, it repeats the optimization with speed/jerk limits 2.2\,m/s and 20\,m/s$^3$ and heights $[0.30,2.00]$\,m, retaining planned separation. Both modes use acceleration feedforward and force limits. Nominal first-interval height, speed and separation are checked before execution.

\begin{figure}[t]
\centering
\includegraphics[width=\columnwidth]{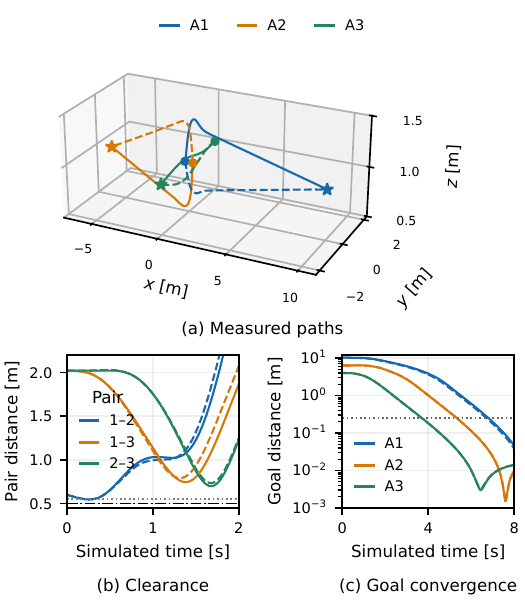}
\caption{Cooperative flights: (a) measured 3D paths, (b) pair distances around the encounter and (c) goal distances. Solid/dashed lines denote GNE/vGNE. Colors identify drones or pairs. Horizontal lines mark physical/planned separation (0.50/0.55\,m) and arrival distance (0.25\,m).}
\label{fig:cooperative_bullet}
\end{figure}

All drones reach their goals in both 30\,s simulations (Figure~\ref{fig:cooperative_bullet}). Arrival requires distance $\leq0.25$\,m and speed $\leq0.30$\,m/s. Joint arrival occurs by 6.81/6.72\,s for GNE/vGNE, with minimum measured pair distances 0.546/0.547\,m. Maximum final position error is 0.041\,m. No sampled physical-limit stop, contact or motor saturation occurs.

All 600 applied plans are model-feasible. Both modes restart once from an improving unilateral response. Numerical checks accept 300/300 GNE and 299/300 vGNE inputs. One vGNE input fails the residual check and executes as a feasible approximation. Thus similar goal reaching coexists with a difference in numerical acceptance. Closed-loop stability and recursive feasibility are not established.

\subsection{Adversarial Pursuit--Evasion}
\label{sec:adversarial}

Three pursuers and two evaders interact with four point obstacles. Pursuers minimize nearest-evader distance, while evaders balance goal tracking against pursuer proximity. A preliminary double-integrator MPC comparison motivates fixed goal/evasion weights 100/10. Pair and obstacle interactions are soft costs with individual hard bounds, giving an ordinary non-zero-sum Nash game. If the solver does not find a feasible plan, it repeats the optimization with heights $[0.30,2.00]$\,m. Other bounds remain unchanged.

At most five sequential response sweeps are used. The update fractions in Section~\ref{sec:method} are $\alpha=0.7$, $0.35$ and $1$, tested in that order. Thus the planner first tests 70\% of the proposed change, then 35\%, and finally the full response if neither smaller update is accepted.

\begin{figure}[t]
\centering
\includegraphics[width=\columnwidth]{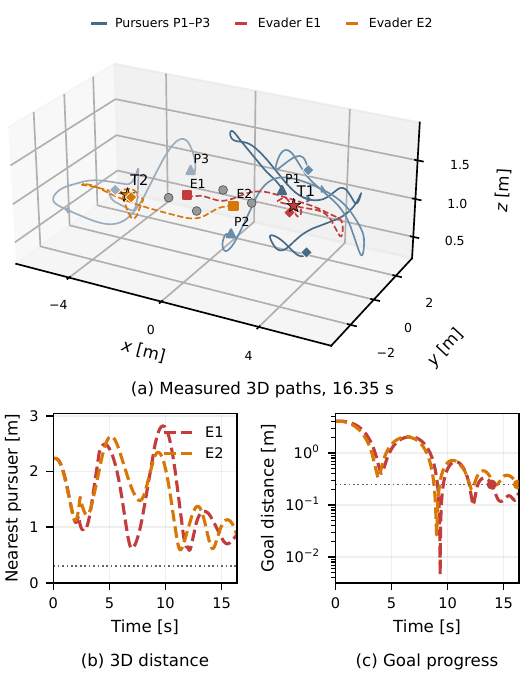}
\caption{Pursuit--evasion: (a) measured 3D paths with pursuers solid and evaders dashed, (b) nearest-pursuer distances and (c) evader goal distances. Stars denote targets and gray dots point obstacles. Horizontal lines mark capture/arrival thresholds. Dots mark individual arrival. The vertical scale is expanded for visibility.}
\label{fig:adv_traj}
\end{figure}

Figure~\ref{fig:adv_traj} shows measured paths, nearest-pursuer distances and goal progress. Both evaders reach their targets without capture. Their minimum nearest-pursuer distances are 0.585/0.593\,m, above the 0.30\,m capture threshold. Both simultaneously satisfy the 0.25\,m distance and 0.30\,m/s speed criteria at 16.35\,s, defining the plotted endpoint. The planner checks returned-plan feasibility without a separate unilateral-response search, so this flight demonstrates task completion rather than verified equilibrium quality.

An otherwise identical PyBullet run using only full response updates ($\alpha=1$) stops at 2.5\,s on a rejected plan. At that state, even maximum upward model acceleration places one drone's next predicted node below the 0.30\,m height limit. This comparison supports retaining damping for the tested planner. It does not establish optimality of the chosen fractions, and the 35\% option is never selected in the damped run.

\subsection{Laboratory Tracking Integration}
\label{sec:hardware}

Five drones track a translating formation using Vicon feedback and Crazyswarm2 \cite{preiss2017crazyswarm}, with off-board computation on an Intel Core i7-8550 laptop. NashOpt's separable linear tracker uses $H=5$, position and input-increment penalties, and $Q_{F,i}=Q_i$. It has no internal input bounds or pair/obstacle constraints. The runner clips commands afterwards, and an external supervisor aborts below 0.50\,m measured separation.

The 56.1\,s flight completes without abort. Across 426 updates, tracking root mean square error is 0.071\,m relative to issued position commands. Maximum complete-loop time is 41.0\,ms, meeting the 100\,ms budget throughout. This validates tracking integration at 10\,Hz, leaving coupled geometric planning untested on hardware.

\section{Conclusion and Future Work}
\label{sec:conclusion}

We developed and evaluated a game-theoretic MPC planner for cooperative and adversarial drone coordination. The solver comparison distinguishes numerical acceptance, equilibrium quality and conflict resolution. Potential minimization resolves the prescribed conflicts more consistently than joint KKT searches under the tested configurations and initializations. Move blocking reduces computation, but computation times for flight-sized coupled problems still exceed the planning period. PyBullet simulations demonstrate cooperative goal reaching and pursuit--evasion, while the laboratory flight supports separable tracking integration.

Further work will investigate feasibility and its preservation across MPC updates, opponent identification, sensing and model uncertainty. Faster numerical methods, including learned warm starts followed by feasibility and optimality refinement \cite{bemporad2026learning}, are needed to approach fixed-rate coupled planning.

\section*{Acknowledgment}

This work was supported by the Technology Agency of the Czech Republic, programme National Competence Centres, project \#TN~0200~0054 Bozek Vehicle Engineering National Competence Center.

\section*{Code Availability}
The companion release provides controllers, experiment configurations, numerical records and commands to verify results and regenerate tables and figures. The repository is available at \url{https://github.com/zboucek/game-theoretic-mpc-drones}.

\bibliographystyle{IEEEtran}
\bibliography{refs}

\end{document}